\documentclass[runningheads]{llncs}

\usepackage[T1]{fontenc}

\usepackage{graphicx}
\usepackage{hyperref}
\usepackage{booktabs, multirow, siunitx, xcolor, etoolbox}

\definecolor{lightgray}{gray}{0.5}

\usepackage{multirow}
\usepackage{subcaption}
\usepackage{listings}
\usepackage{graphicx}
\usepackage{tikz}
\usepackage{subcaption}
\usepackage{pgfplots}
\usepackage{enumitem}
\usepackage{orcidlink}
\usepackage{amsfonts}
\usepackage{amsmath}
\usepackage{mathtools}
\usepackage{amssymb}
\usepackage{rotating}
\usepackage{bbm}
\usepackage{minted}
\definecolor{sdgray}{gray}{0.6}

\definecolor{todocolor}{rgb}{0.9,0.1,0.1}
\definecolor{changedcolor}{rgb}{0.42,0.27,0.57}
\definecolor{addedcolor}{rgb}{0.867,0.176,0.361}

\def\TraUniverse{{\ensuremath{\Theta}}}
\def\ActUniverse{{\ensuremath{\Sigma}}}
\def\NodeUniverse{{\ensuremath{\mathcal{N}}}}
\def\lab{\ensuremath{\ell}}

\definecolor{darkgreen}{cmyk}{1, 0, 1, 0.5}

\setlist{topsep=2pt,itemsep=1pt,parsep=0pt,partopsep=0pt}

\begin{document}
\title{Beyond Control-Flow: Integrating the Resource Perspective into Multi-Collaborative Process Modeling from Text\thanks{Preprint version submitted to EDOC 2026. This version
has not undergone peer review}}

\titlerunning{Resource-Aware Multi-Collaborative Process Modeling from Text}

\author{Anton Antonov\inst{1,2}\orcidID{0009-0004-1044-4884} \and
Humam Kourani\inst{1,2}\orcidID{0000-0003-2375-2152} \and
Alessandro Berti\inst{2}\orcidID{0000-0002-3279-4795} \and Gyunam Park\inst{1}\orcidID{0000-0001-9394-6513}
}

\authorrunning{A. Antonov et al. }

\institute{Fraunhofer Institute for Applied Information Technology FIT, Schloss Birlinghoven, 53757 Sankt Augustin, Germany
\email{\{anton.antonov,humam.kourani,gyunam.park\}@fit.fraunhofer.de} \and
RWTH Aachen University, Ahornstraße 55, 52074 Aachen, Germany\\
\email{a.berti@pads.rwth-aachen.de}
}

\maketitle
\begin{abstract}
Process modeling is a sub-domain of Business Process Management (BPM) focused on the translation of process artifacts into formal models. This task traditionally requires extensive human input and domain expertise in both BPM notations and the specific business context. While Large Language Models (LLMs) can now automate much of this manual work, current text-to-model approaches focus predominantly on the control-flow perspective—ordering activities without considering the collaborative aspect of the processes. In this paper, we introduce a resource-aware generation pipeline that produces formal BPMN 2.0 collaboration diagrams from natural-language descriptions. Rather than solely prompting an LLM for raw XML, we describe a compact, executable intermediate language with mandatory resource details defining both the organization (pool) and the role (lane). Cross-organization dependencies are materialized using the standard formal notation for such interactions—message events—while an orthogonal layout routine automatically handles the spatial arrangement of elements within pools and lanes. Experiments on ten business processes with nine LLMs show strong resource discovery while preserving control-flow quality and adding only marginal runtime overhead. This approach moves generative modeling toward a more comprehensive, multi-collaborative representation of business operations.
\end{abstract}

\section{Introduction}
\label{sec:introduction}

\begin{figure}[t!]
    \centering
    
    \begin{minipage}[b]{\textwidth}
        \centering
        \includegraphics[width=0.95\textwidth]{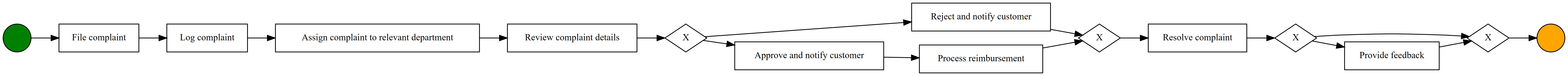} 
        \par\vspace{4pt}
        {\small (a) Baseline generation (ProMoAI): The model captures the logical sequence using standard control-flow elements: \emph{tasks}, \emph{start/end events}, \emph{gateways}, and \emph{sequence flows}. However, the organizational context is missing.}
    \end{minipage}

    \begin{minipage}[b]{\textwidth}
        \centering

        \includegraphics[width=0.95\textwidth]{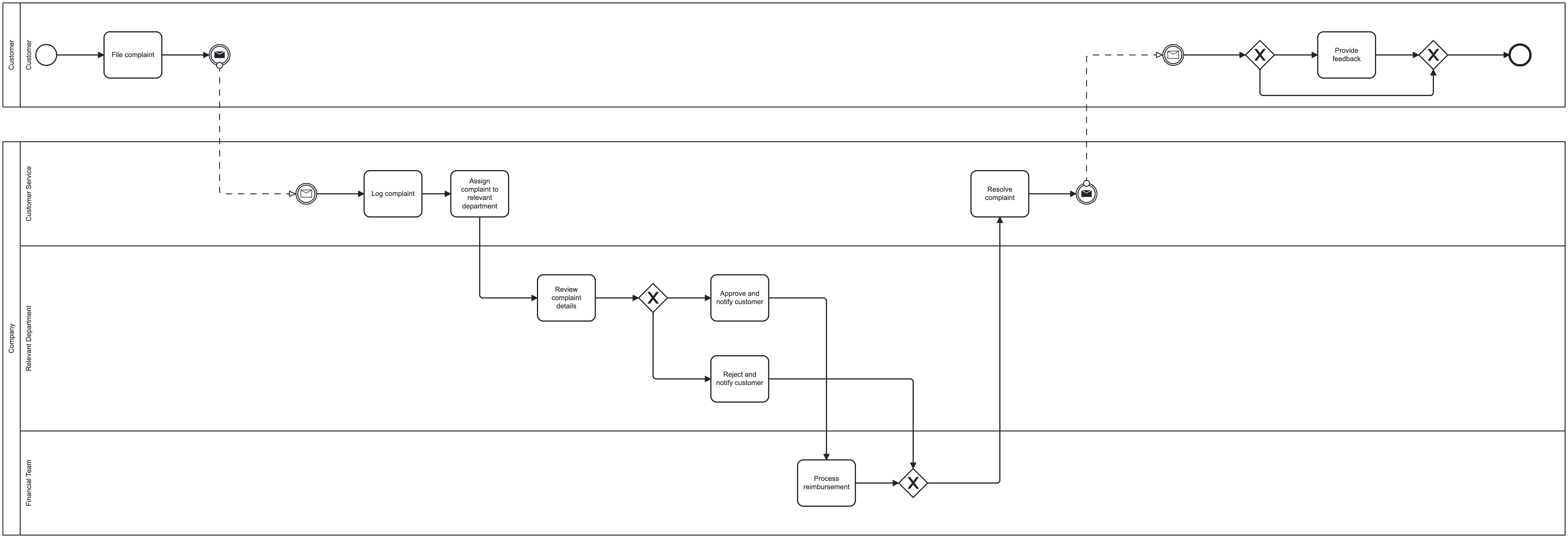}
        \par\vspace{4pt}
        {\small (b) Our approach: The model integrates the organizational perspective. In addition to control flow, it utilizes \emph{pools} and \emph{lanes} to define roles, and \emph{message flows} with \emph{message events} to represent inter-organizational communication.}
    \end{minipage}
    
    \caption{Motivating Example: Models generated by \emph{gemini-3-flash} for the \textit{Complaint Handling} process (p13).}
    \label{fig:motivating_example}
\end{figure}

Enterprise processes are inherently \emph{collaborative}. They involve
organizational units, roles, systems, customers, suppliers, and other
participants whose interactions shape how work is coordinated across
organizational boundaries. A process model therefore needs more than the
\emph{control-flow perspective}, i.e., the ordering of activities and events:
it must also capture the \emph{resource perspective}, namely who performs each
activity and how participants communicate. This organizational view is
essential for understanding responsibilities, handovers, and operational
interfaces between collaborating actors.
To illustrate the importance of the resource perspective, consider the \textit{Complaint Handling} process depicted in \autoref{fig:motivating_example}. Unlike the ``flat'' model generated using \cite{DBLP:conf/bpmds/KouraniB0A24} in \autoref{fig:motivating_example}(a) which lacks departmental distinctions, our proposed method in (b) explicitly models resources through pools and lanes. This enrichment transforms a simple sequence of steps into a comprehensive blueprint of organizational interaction and communication.

Business Process Model and Notation (BPMN) 2.0 supports such multi-perspective
modeling through \emph{pools}, \emph{lanes}, and \emph{message
flows}~\cite{omg2011bpmn}. Yet constructing BPMN collaboration models remains
challenging, especially for non-experts, as it requires both domain knowledge
and familiarity with the notation's syntax and semantics
\cite{DBLP:journals/bpmj/Recker10,DBLP:journals/bpmj/Rosemann06a}. Recent
LLM-based approaches reduce manual effort by generating models from
natural-language descriptions
\cite{DBLP:conf/bpm/KopkeS24,DBLP:conf/bpmds/KouraniB0A24}, but predominantly
focus on control flow, leaving the organizational structure of the process
underspecified.

Adding the resource perspective is not a simple annotation task.
Organizational information is often implicit in text, and activity ordering,
responsibility assignment, and cross-participant communication must be
inferred consistently. Moreover, directly generating BPMN XML forces the LLM to
juggle process logic, resource assignment, syntactic correctness, and
serialization simultaneously, making the output error-prone and hard to
validate. A more robust solution is to separate \emph{model generation} from
\emph{formal serialization} while preserving the semantic coupling between
behavior and resources.

\begin{figure}[!t]
    \centering
      \includegraphics[width=\textwidth]{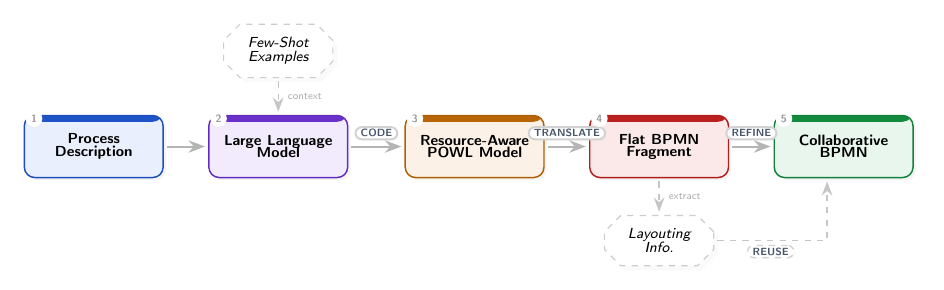}
\caption{Overview of the proposed resource-aware generation pipeline: from natural-language descriptions via a compact intermediate representation to BPMN collaboration models with automated layouting.}
    \label{fig:paper_outline}
\end{figure}

We address this by extending a grammar-guided process generation pipeline \cite{DBLP:conf/ijcai/KouraniB0A24} with a \emph{resource-aware intermediate representation}. In this representation, each generated activity is explicitly associated with a \emph{pool} and a \emph{lane}, so that control-flow structure and resource assignments are produced jointly rather than in separate
post-processing steps. The result is transformed deterministically into a standard BPMN 2.0 collaboration model: intra-participant dependencies are kept as \emph{sequence flows}, and cross-participant dependencies are represented through \emph{message events} and \emph{message flows}. An automated \emph{layouting} routine then arranges the model into pools and lanes.
\autoref{fig:paper_outline} summarizes the overall approach.

The paper makes three contributions: (i) a \emph{resource-aware intermediate
modeling language} for generating multi-perspective process models from text;
(ii) a \emph{deterministic transformation} from this representation to
standard BPMN collaboration diagrams; and (iii) an evaluation across multiple
business processes and LLMs showing that the added resource perspective
preserves control-flow quality while enabling accurate organizational
discovery with only marginal overhead.

The remainder of the paper is organized as follows. \autoref{chap:related_work}
reviews related work. \autoref{chap:background} introduces BPMN and the
intermediate representation we build on. \autoref{chap:method} presents the
resource-aware extension, the low-level modeling language, the transformation
to BPMN, and the layouting procedure. \autoref{chap:eval} reports the
evaluation, and \autoref{chap:conclusion} concludes.

\section{Related Work}
\label{chap:related_work}

We position our work along two lines of research: automated process modeling and multi-perspective process modeling.

\paragraph{Automated Process Modeling}
We build on the grammar-guided process generation paradigm proposed in \cite{DBLP:conf/ijcai/KouraniB0A24,DBLP:conf/bpmds/KouraniB0A241}, which ensures structural correctness and soundness by design, keeping the generated control flow valid as model complexity grows. We extend this foundation beyond the control-flow perspective by introducing a resource-aware intermediate representation that natively supports organizational constructs. Other GenAI approaches for process modeling face distinct limitations. Some rely on resource-intensive model fine-tuning \cite{DBLP:conf/icsoc/NivonS24}, while others adopt generic data formats such as JSON \cite{DBLP:conf/bpm/KopkeS24,DBLP:journals/corr/abs-2509-24592}. Being generic, JSON is not optimized for process semantics and struggles to express patterns like nested gateways or loops of choices consistently. A different line of work \cite{DBLP:journals/corr/abs-2408-01916} proposes a multi-agent framework that emulates the iterative stages of manual process modeling; while accurate, its complexity introduces significant computational overhead and makes a human-in-the-loop infeasible, since injecting user feedback into such a convoluted workflow is non-trivial.

\paragraph{Multi-Perspective Process Modeling}
A limited number of studies have addressed multi-perspective modeling. For instance, \cite{da00583d626a44e89033f1589296753f} discuss a method for integrating the data perspective. However, their implementation relies on internal structures and transformation steps that are not fully disclosed. Additionally, their approach produces models that depart from the standard BPMN specification, reducing their compatibility with existing process management tools. While \cite{safan2025bpmn} extend current approaches to include both resource and data perspectives, their contribution prioritizes engineering specifics over a formal definition of the underlying method, leaving critical transformation steps undocumented. Our work addresses these gaps by providing a transparent, deterministic pathway from natural language to standard-compliant, multi-perspective models.

\section{Background}
\label{chap:background}

Before presenting our approach, we introduce the BPMN 2.0 notation used as the target formalism and the POWL intermediate representation.

\subsection{Business Process Model and Notation 2.0}
BPMN 2.0 \cite{omg2011bpmn} is the industry standard for multi-perspective process modeling, providing a formal metamodel for representing different business dimensions. Structurally, a BPMN model is composed of nodes, including tasks, events, and gateways, and edges, such as sequence flows and message flows. These elements capture mostly the control-flow perspective, while pools and lanes provide support for representing the organizational perspective.

Within this hierarchy, \textit{pools} represent distinct organizational participants, while \textit{lanes} partition pools into specific roles or departments. A fundamental formal constraint is the separation of internal logic from inter-organizational interaction: communication across pool boundaries must be modeled via \textit{Message Flows}, whereas internal coordination relies exclusively on \textit{Sequence Flows}.

\subsection{Partially-Ordered Workflow Language, ProMoAI}
The Partially Ordered Workflow Language \cite{DBLP:conf/bpm/KouraniPA25} (POWL) is a hierarchical intermediate representation that ensures structural soundness by construction. POWL models are defined recursively using activity instances as a foundational building block that are further grouped into choice graphs and partial orders. 
Let the set of all activity labels be denoted with $\Sigma$. We use $\tau \notin \Sigma$ for a silent activity. To enable creating process models with duplicated activities, we use the notion of \emph{transitions}. Let $\TraUniverse$ denote the universe of transitions. Each transition is mapped to an activity label by the labeling function $\lab:\TraUniverse\to\ActUniverse\cup\{\tau\}$. A transition $t$ is \emph{visible} if $\lab(t)\neq\tau$ and silent otherwise.

 A strict partial order on a set $X$ is a binary relation $\prec \ \subseteq X \times X$ such that for all $x,y,z \in X$, $x \nprec x\text{ (irreflexive) and }(x \prec y \land y \prec z) \Rightarrow x \prec z
$ (transitive). For a partial order $\prec$, let $\prec^-$ denote its transitive reduction, defined by $x \prec^- y
\iff
x \prec y
\ \text{and there is no } z \text{ with } x \prec z \land z \prec y$. A choice graph over a set of nodes $X$ is a directed graph
$G=(V,E)$ with $V = X \cup \{\vartriangleright,\square\}$, where
$\vartriangleright,\square \notin X$, such that $\vartriangleright$ is the unique
source, $\square$ is the unique sink, and every node lies on a directed path
from $\vartriangleright$ to $\square$.

Semantically, choice graphs state that exactly one path from start to end is executable. For a partial order, it models all interleavings of the submodel languages that respect the specified ordering constraints. 

ProMoAI \cite{DBLP:conf/ijcai/KouraniB0A24} is a framework that turns natural language descriptions of processes to formal process models based on POWL.  ProMoAI's architecture \cite{DBLP:conf/ijcai/KouraniB0A24} comprises several components: low-level intermediate representation, prompting module, process generation, error handling, and user refinement.

\section{Method}
\label{chap:method}

We first extend POWL\footnote{\url{https://github.com/fit-process-mining/POWL}} and the ProMoAI generation pipeline\footnote{\url{https://github.com/fit-process-mining/ProMoAI}} with an explicit resource perspective. Second, we define a deterministic transformation from resource-aware POWL models into valid BPMN 2.0 collaboration diagrams. Third, we describe an automated orthogonal layouting procedure for multi-perspective BPMN models.

\subsection{Extending ProMoAI to Resource-Aware POWL Generation}
To incorporate the resource perspective, we refine POWL over resource-aware activity transitions. Instead of changing the recursive structure of POWL, we enrich each observable activity with the organizational context in which it is executed, namely a pool and a lane.


\begin{definition}[Organizational Assignment]
Let $\mathcal{P}$ be the set of pools. For each pool $p \in \mathcal{P}$, let $L_p$ be the set of lanes contained within that pool. We define the set of all valid organizational contexts as:
$$\mathcal{C} = \{(p, l) \mid p \in \mathcal{P}, l \in L_p\}$$
The resource assignment function is the total function $\Lambda: \{t \mid t \in \TraUniverse, \lab(t) \not = \tau\} \rightarrow \mathcal{C}$, which maps each visible transition $t$ to a specific pool-lane pair $(p, l) \in \mathcal{C}$.
\end{definition}

As evident from the definition, a lane with the same name can exist in different pools. This reflects the fact that distinct organizations modeled within a single collaboration may share the same conceptual roles (e.g., ``Manager'' or ``Sales''). By scoping lanes to their respective pools, our formalization ensures that these roles are uniquely identified through their organizational context, even when they share identical labels in the process narrative.

We use this added perspective to extend POWL: each observable transition $t$ is represented together with its resource context $\Lambda(t)$, while silent transitions remain unassigned because they capture internal control-flow logic rather than executable work. Accordingly, we adapt the low-level modeling language used in ProMoAI and revise the prompt engineering, few-shot examples, model generation, validation, and error-handling steps so that the pipeline produces resource-aware POWL models.

\subsection{Generating Resource-Aware BPMN Models}
Transforming POWL into BPMN follows a recursive approach, leveraging the fact that POWL models are inherently recursive. In ProMoAI \cite{DBLP:conf/bpmds/KouraniB0A24} this conversion relied on an intermediate Petri net representation. However, standard Petri nets do not natively support organizational constructs (pools, lanes, message events, and message flows). To address this, we introduce a direct translation pipeline (POWL $\rightarrow$ BPMN).

The transformation from resource-aware POWL to BPMN proceeds in three steps. First, we translate the POWL control-flow structure into a flat BPMN fragment while preserving the semantics of activities, partial orders, and choice graphs. Second, we lift the transition-level resource assignments to BPMN flow nodes by assigning each node to a pool and lane. Third, since BPMN does not permit sequence flows across pool boundaries, we replace cross-pool dependencies with explicit message events and message flows, yielding a valid BPMN 2.0 collaboration diagram.

\noindent \textbf{Step 1: Control-Flow Translation}

To enable the control-flow translation, we first define \emph{BPMN fragments} as the target structure of the recursive translation. 

Let $\NodeUniverse$ denote the universe of BPMN flow nodes. We assume a fixed set
of BPMN node types
\[
\mathbf{Types}
=
\{
\mathsf{task},
\mathsf{startEvent},
\mathsf{endEvent},
\mathsf{andGateway},
\mathsf{xorGateway},
\]
\[\mathsf{throwMsg}, \mathsf{catchMsg}
\}.
\]
The BPMN type assignment is given by the total function
\[
\kappa : \NodeUniverse \rightarrow \mathbf{Types},
\]
which assigns each BPMN node its node type. Activity labels are assigned by the
partial function
\[
\Gamma : \NodeUniverse \rightharpoonup \Sigma,
\]
where $\Gamma(n)$ is defined iff $\kappa(n)=\mathsf{task}$.

\begin{definition}[BPMN fragment]
A BPMN fragment is a tuple $F = (N, \allowbreak S, \allowbreak s, \allowbreak e)$, where $N \subseteq \NodeUniverse$ is a finite
set of nodes; $S \subseteq N \times N$ is the set of sequence flows; and $s, e \in N$ are the auxiliary start and end connector nodes used to compose fragments.
\end{definition}
The translation deterministically maps a resource-aware POWL model to a BPMN fragment, where transitions are mapped into tasks, partial orders into concurrency structures, and choice graphs into exclusive choice structures.

\begin{definition}[Control-flow translation]
    The translation $\mathcal{T}$ maps a resource-aware POWL model $\psi$ to a BPMN fragment, i.e., $\mathcal{T}(\psi) = (N, S, s, e)$, as follows. Throughout all cases below, $s$ and $e$ denote the unique entry and exit nodes of the constructed fragment, respectively, and their BPMN types are fixed as
    $\kappa(s)=\mathsf{startEvent}$ and $\kappa(e)=\mathsf{endEvent}$.
        \begin{enumerate}[label=(\roman*)]
    
    \item \textbf{Visible transitions.} A visible transition $t \in \TraUniverse$ with $\lab(t) \neq \tau$ is translated into a task node $n_{t}$ and two auxiliary nodes $s, e$:
    \[N = \{s, n_{t}, e\}, S = \{(s, n_{t}), (n_{t}, e)\}, \Gamma(n_{t})=\lab(t), \kappa(n_{t})=\mathsf{task}.\]
    \item \textbf{Silent transitions.} A silent transition $t \in \TraUniverse$ with $\lab(t) = \tau$ is translated into two auxiliary nodes $s, e$:
    \[N = \{s, e\}, S = \{(s, e)\}.\]
\item \textbf{Partial orders.} Let $\prec$ be a partial order over a set of POWL models $\{\psi_1, \ldots, \psi_n\}$. We introduce a
global parallel split $g_{\land}^{+}$, i.e., $\kappa(g_{\land}^{+})=\mathsf{andGateway}$, a global parallel join $g_{\land}^{-}$, i.e. $\kappa(g_{\land}^{-})=\mathsf{andGateway}$,  and, for
every submodel $\psi_i$, a dedicated parallel join $g_{\land i}^{-}$ and a dedicated parallel split
$g_{\land i}^{+}$, i.e., $\kappa(g_{\land i}^{+})=\mathsf{andGateway},\kappa(g_{\land i}^{-})=\mathsf{andGateway}$ for each $i$. Let $\mathcal{T}(\psi_i) = (N_i, S_i, s_i, e_i)$ be the translation
of $\psi_i$. The node set is defined as
\[
N =
\{s,e,g_{\land}^{+},g_{\land}^{-}\}
\cup
\{g_{\land i}^{+},g_{\land i}^{-} \mid 1 \leq i \leq n\}
\cup
\{n \mid n \in N_i \text{ for some } 1 \leq i \leq n\}.
\]

The sequence-flow relation $S$ consists of the following edges:
\[
\begin{array}{ll}
(s, g_{\land}^{+}), (g_{\land}^{-}, e), &  \\[2pt]
(g_{\land i}^{-}, s_i),\ (e_i, g_{\land i}^{+}) \;\;\; & \text{for every } \psi_i, \\[2pt]
(g_{\land}^{+}, g_{\land i}^{-}) & \text{for every $\psi_i$ s.t. $\not\exists \psi': \psi' \prec \psi_i$}, \\[2pt]
(g_{\land i}^{+}, g_{\land}^{-}) & \text{for every $\psi_i$ s.t. $\not\exists \psi': \psi_i \prec \psi'$}, \\[2pt]
(g_{\land i}^{+}, g_{\land i'}^{-}) & \text{for every } \psi_i \prec^{-} \psi_{i'}.
\end{array}
\]

\item \textbf{Choice graphs.} Let $G = (V,E)$ be a choice graph over POWL models $\psi_1, \ldots, \psi_n$ with edge set $E$. We introduce a global exclusive split $g_{\times}^{+}$, a global
exclusive join $g_{\times}^{-}$, and, for every submodel $\psi_i$, an exclusive join
$g_{\times i}^{\mathrm{-}}$ placed before $s_i$ and an exclusive split
$g_{\times i}^{\mathrm{+}}$ placed after $e_i$. Let
$\mathcal{T}(\psi_i) = (N_i, S_i, s_i, e_i)$ be the translation of $\psi_i$. The node set is defined as
\[
N =
\{s,e,g_{\times}^{+},g_{\times}^{-}\}
\cup
\{g_{\times i}^{+},g_{\times i}^{-} \mid 1 \leq i \leq n\}
\cup
\{n \mid n \in N_i \text{ for some } 1 \leq i \leq n\}.
\]
The sequence-flow relation $S$ consists of the following edges:
\[
\begin{array}{ll}
(s, g_{\times}^{+}), (g_{\times}^{-}, e), &  \\[2pt]
(g_{\times i}^{-}, s_i),\ (e_i, g_{\times i}^{+}) \;\;\; & \text{for every } \psi_i, \\[2pt]
(g_{\times}^{+}, g_{\times i}^{-}) & \text{for every $\psi_i$ s.t. $(\vartriangleright, \psi_i) \in E$}, \\[2pt]
(g_{\times i}^{+}, g_{\times}^{-}) & \text{for every $\psi_i$ s.t. $(\psi_i, \square) \in E$}, \\[2pt]
(g_{\times i}^{+}, g_{\times i'}^{-}) & \text{for every } (\psi_i, \psi_{i'}) \in E.
\end{array}
\]


\end{enumerate}\end{definition}
Gateways with a single incoming and outgoing flow are semantically transparent
and may be removed in a trivial post-processing step; they are retained here
for uniformity. The auxiliary start and end events for submodels are likewise redundant in the final output and are pruned.

\noindent \textbf{Step 2: Pool and Lane Assignment}

After the control-flow translation, organizational information is available only for task nodes, since $\Lambda$ is defined on observable transitions. BPMN collaborations, however, require each flow node to be placed inside a pool, and in our setting inside a concrete lane of that pool. We therefore lift the transition-level assignment to a node-level organizational assignment.

Let $F = (N,S,s,e)$ be the translated BPMN fragment. For any visible transition $t\in \TraUniverse, \lab(t) \not = \tau$, let $n_t \in N$ denote its mapped task node in $F$. We define a partial assignment
$\chi_0 : N \rightharpoonup \mathcal{C}$ that assigns each task node to the organizational context of its corresponding transition:
\[
\chi_0(n_t) = (p,l)
\quad\text{iff}\quad
\kappa(n_t)=\mathsf{task} \quad \text{and} \quad \Lambda(t)=(p,l).
\]

The assignment is then propagated recursively to the remaining flow nodes. Let
$\mathrm{pred}(n)=\{m \in N \mid (m,n)\in S\}$ and
$\mathrm{succ}(n)=\{m \in N \mid (n,m)\in S\}$
denote the direct predecessors and successors of a node $n$. Starting from $\chi_0$, we repeatedly assign each unassigned node to the organizational context of an already assigned neighboring node. The relevant neighborhood depends on the node type: for split gateways, we consider only successors, since a split routes control flow into the following branches; for join gateways, we consider only predecessors, since a join synchronizes completed incoming branches; for start and end events, we consider the adjacent assigned node in the control-flow direction. This propagation continues until all gateways and boundary events are assigned. In cases where several neighboring contexts are available at the same distance, the assignment is not semantically unique; any of these contexts yields a valid BPMN placement, and the implementation resolves such ties deterministically. The result is a total node-level organizational assignment $\chi : N \rightarrow \mathcal{C}$.


\noindent \textbf{Step 3: Adding Message Events and Message Flows}

Once all BPMN nodes have been assigned to pools and lanes, we enforce the BPMN collaboration constraint that sequence flows may only connect nodes within the same pool. Dependencies between different pools are therefore materialized as message-based interactions: each cross-pool sequence flow is replaced by a throwing message event in the source pool, a catching message event in the target pool, and a message flow connecting both events.

\begin{definition}[BPMN collaboration skeleton]
A BPMN collaboration skeleton is a tuple
\[
CS = (P,\{L_p\}_{p \in P},\{N_p\}_{p \in P},\{S_p\}_{p \in P},M,\lambda_\mathcal{C}),
\]
where $P \subseteq \mathcal{P}$ is the set of pools, $L_p \subseteq \mathcal{L}_p$
is the set of lanes of pool $p$, $N_p$ is the set of flow nodes located in pool $p$,
$S_p \subseteq N_p \times N_p$ is the participant-local sequence-flow relation,
$M \subseteq \{(u,v) \mid u \in N_p,\ v \in N_q,\ p,q \in P,\ p \neq q\}$ is the set of message flows,
$\lambda_\mathcal{C} : \bigcup_{p \in P} N_p \rightarrow \mathcal{C}$ assigns each node to its pool-lane pair. 
Furthermore, the node sets are disjoint, i.e., $N_p \cap N_{p'}=\emptyset$ for all $p, p' \in \mathcal{P}$.
\end{definition}

Let $F=(N,S,s,e)$ be the flat BPMN fragment obtained in Step~1 after pruning auxiliary nodes, and let $\chi:N\rightarrow\mathcal{C}$ be the total node-level organizational assignment from Step~2. For convenience, we write $\mathrm{pool}(n)=p
\quad\text{and}\quad
\mathrm{lane}(n)=l
\quad\text{iff}\quad
\chi(n)=(p,l)$. The cross-pool sequence flows are $\mathcal{X}=\{(u,v)\in S \mid \mathrm{pool}(u)\neq \mathrm{pool}(v)\}$.

For every $(u,v)\in\mathcal{X}$, we introduce a throwing message event $m_{uv}^{+} \in \NodeUniverse$ and a catching message event $m_{uv}^{-} \in \NodeUniverse$:
\[
\kappa(m_{uv}^{+})=\mathsf{throwMsg}
\qquad \text{and} \qquad
\kappa(m_{uv}^{-})=\mathsf{catchMsg}.
\]

Let $N^{+}
=
N
\cup
\{m_{uv}^{+},m_{uv}^{-} \mid (u,v)\in\mathcal{X}\}$ be the node set after inserting message events.

The collaboration skeleton $C = (P,\{L_p\}_{p \in P},\{N_p\}_{p \in P},\{S_p\}_{p \in P},M,\lambda_C)$ is then derived as follows:
\begin{itemize}
    \item $P = \{\mathrm{pool}(n) \mid n \in N^{+}\},$

\item And for every $p\in P$:
\begin{itemize}
    \item $L_p = \{\mathrm{lane}(n) \mid n \in N^{+},\ \mathrm{pool}(n)=p\}$,
    \item $N_p = \{n \in N^{+} \mid \mathrm{pool}(n)=p\},$
    \item $S_p = \{(u,v) \in S \setminus \mathcal{X} \mid u,v \in N_p\} \quad \cup \quad \{(u,m^{+}_{uv}) \mid (u,v) \in \mathcal{X},\ u \in N_p\} \quad \cup \quad \{(m^{-}_{uv},v) \mid (u,v) \in \mathcal{X},\ v \in N_p\}$
\end{itemize}

\item $M = \{(m^{+}_{uv},m^{-}_{uv}) \mid (u,v) \in \mathcal{X}\}$.
 \item The organizational assignment is defined as follows: 
\[
\lambda_\mathcal{C}(n)=\chi(n)\quad\text{for }n\in N,
\]
\[
\text{and }\lambda_\mathcal{C}(m_{uv}^{+})=\chi(u),
\quad
\lambda_\mathcal{C}(m_{uv}^{-})=\chi(v) \quad\text{for }(u,v) \in \mathcal{X}. 
\]

\end{itemize}


\subsection{Collaboration Layout Generation}
\label{ssec:layout}
Finally, we generate diagram interchange information for the BPMN collaboration
skeleton. The layout is constructed by the following deterministic steps:
\begin{enumerate}[label=\textbf{L\arabic*.}, leftmargin=3.2em]
    \item \textbf{Initial node positioning.} The layout procedure starts from a flat graph layout \cite{DBLP:conf/gd/Sander95} that considers only the flow nodes and their connections, without taking the resource constraints into account. We denote this initial visual placement of a node $n$ by $\rho_0(n)$. The exact representation of $\rho_0(n)$ may include coordinates and dimensions, such as position, width, and height.

    \item \textbf{Pool ordering.} Pools are placed from top to bottom according
    to the leftmost node assigned to each pool in the initial layout $\rho_0$.

    \item \textbf{Lane ordering.} Within each pool, lanes are placed from top to
    bottom according to the leftmost node assigned to each lane in $\rho_0$.

    \item \textbf{Node placement.} Each node is shifted into the rectangle of its assigned pool and lane. The horizontal order induced by $\rho_0$ is preserved, while only the vertical position is updated.

    \item \textbf{Edge routing.} After all flow nodes have been placed, the waypoints of each edge are recomputed using an orthogonal,
    obstacle-aware routing heuristic \cite{DBLP:conf/gd/Sander95}.
    
\end{enumerate}

\section{Evaluation}
\label{chap:eval}
In this section, we evaluate our approach against the baseline ProMoAI \cite{DBLP:conf/bpmds/KouraniB0A24}, serving as a resource-agnostic benchmark for task complexity, using a diverse set of nine state-of-the-art LLMs. Our evaluation is guided by the following research questions:
\begin{enumerate}[label=\textbf{RQ\arabic*}, leftmargin=3.5em]
    \item \textbf{(Impact of Task Complexity):} Does the requirement to model the resource perspective simultaneously with the control flow degrade the behavioral quality of the resulting process logic?
    
    \item \textbf{(Semantic Discovery Quality):} How accurately can LLMs identify organizational entities (pools) and functional roles (lanes) compared to human-verified ground truth?
    
    \item \textbf{(Operational Efficiency):} What is the overhead of multi-perspective generation in terms of runtime performance and the stability of the self-correction loop?
\end{enumerate}
\subsection{Experimental Settings}
We evaluate our approach on a dataset of ten diverse business processes selected from the benchmark presented in \cite{DBLP:conf/bpmds/KouraniB0A24}. We filtered the original dataset to retain only processes that contain sufficient detail regarding process participants. Since the original ground truth BPMNs were limited to control flow, we manually annotated them with pools and lanes. This curation was necessary as existing large repositories of process models (e.g., \cite{DBLP:conf/bpm/Corradini0P0021,DBLP:conf/icpm/SolaWSBRK22}) either lack paired textual descriptions or the ground-truth organizational layers necessary to evaluate multi-perspective modeling. \autoref{tab:selectedprocesses} summarizes the characteristics of the selected processes.

\begin{table*}[!t]
    \caption{Characteristics of the selected ground truth processes.}
    \label{tab:selectedprocesses}
    \centering 
    \resizebox{\textwidth}{!}{
    
    \begin{tabular}{lccccccc}
        \toprule
        \multirow{2}{*}{\textbf{Process}} & \multicolumn{4}{c}{\textbf{Model Metrics}} & \multicolumn{3}{c}{\textbf{Structural Features}} \\
        \cmidrule(lr){2-5} \cmidrule(lr){6-8}
         & \# Act. & \# Gateways & \# Pools & \# Lanes & Decis. & Cycles & Concur. \\
        \midrule
        Sales Order (p1)         & 8  & 6 & 2 & 4 & \checkmark & & \checkmark \\
        Hiring Process (p2)         & 16  & 10 & 1 & 3 & \checkmark & \checkmark & \checkmark \\
        Procurement Process (p5)         & 11  & 4 & 2 & 2 & \checkmark & \checkmark & \checkmark \\
        Booking System (p7)      & 13 & 20 & 2 & 3 & \checkmark & \checkmark & \checkmark \\
        Incident Reporting (p8)      & 13 & 4 & 2 & 3 & \checkmark &  & \checkmark \\
        Prototype Building (p9)      & 13 & 8 & 1 & 2 & \checkmark & \checkmark & \checkmark \\
        Subscription Service (p11) & 12  & 12 & 2 & 3 & \checkmark & \checkmark & \checkmark \\
        Complaint Handling (p13) & 9  & 4 & 2 & 4 & \checkmark & & \\
        Internal Audit (p16)     & 24 & 14 & 1 & 5 & \checkmark & \checkmark & \checkmark \\
        University Admission (p18)     & 26 & 22 & 2 & 6 & \checkmark & \checkmark & \checkmark \\

        \bottomrule
    \end{tabular}
    }
\end{table*}

To demonstrate the model-agnostic nature of our approach, we perform the evaluation using a diverse set of nine LLMs. This selection includes proprietary state-of-the-art models (e.g., from OpenAI, Anthropic, and Google) as well as open-source models\footnote{All evaluation artifacts are available at \url{https://github.com/fit-process-mining/resource-perspective-gen-pm}}.

\subsection{Evaluation Metrics}
\label{ssec:evaluation}
\subsubsection{(i) Effect of Added Task Complexity}
A primary concern in extending generative modeling is whether the increased cognitive load (requiring the LLM to manage pools and lanes simultaneously with control flow) negatively impacts the generation process. We compare our resource-aware approach against the baseline (ProMoAI without resources) using the following metrics:
\begin{description}
    \item[Control-flow Quality Score:]  We report the control-flow quality as the $F_1$-score of process-mining fitness and precision, $F_1 = 2 \cdot \frac{\mathit{fitness}\cdot\mathit{precision}}{\mathit{fitness}+\mathit{precision}}$, computed as described in \cite{DBLP:conf/bpmds/KouraniB0A24}.This metric quantifies how well the behavioral semantics of the generated model align with the ground-truth event logs. A comparable score between the baseline and our method would indicate that adding resource constraints does not disrupt the logic of the process.
    
    \item[Number of Needed Iterations:] Since our pipeline includes an automated self-correction mechanism, we track the number of iterations required to produce a syntactically valid and executable model. An increase in iterations would suggest that the model struggles to adhere to the extended syntax or logical constraints.
    
    \item[Time per Iteration:] Beyond the number of errors, we analyze the execution time per generation step. This metric helps distinguish whether the added complexity forces the LLM to spend more time ``thinking'' or generating longer token sequences.
\end{description}

\begin{table}[!b]
    \centering
        \centering
        \renewcommand{\arraystretch}{1.2}
            \caption{Control-flow quality scores (Mean $\pm$ SD): baseline vs.\ resource-aware. Paired t-test \textit{p}-values show no significant difference ($p>0.05$).}
        \label{tab:f1_stats}
        \begin{tabular}{lccc}
            \toprule
            \multirow{2}{*}{\textbf{Model}} & \multicolumn{2}{c}{\textbf{$F_1$}} & \multirow{2}{*}{\textbf{\textit{p}-value}} \\
            \cmidrule(lr){2-3}
             & \textbf{Baseline} & \textbf{Ours} & \\
            \midrule
            Grok-4 Fast Reasoning & $0.921 \pm 0.072$ & $\mathbf{0.924} \pm 0.069$ & $0.704$ \\
            DeepSeek-v3.2         & $\mathbf{0.917} \pm 0.041$ & $0.895 \pm 0.062$ & $0.231$ \\
            GPT 5.2               & $0.909 \pm 0.075$ & $\mathbf{0.915} \pm 0.064$ & $0.690$ \\
            Gemini-3 Flash        & $0.910 \pm 0.072$ & $\mathbf{0.920} \pm 0.058$ & $0.452$ \\
            Claude Sonnet 4.5     & $\mathbf{0.916} \pm 0.070$ & $0.899 \pm 0.069$ & $0.522$ \\
            Kimi K2               & $\mathbf{0.895} \pm 0.067$ & $0.835 \pm 0.118$ & $0.103$ \\
            Claude Haiku 4.5      & $0.876 \pm 0.076$ & $\mathbf{0.902} \pm 0.066$ & $0.172$ \\
            Qwen3 Next 80B        & $\mathbf{0.876} \pm 0.073$ & $0.863 \pm 0.093$ & $0.575$ \\
            GPT-5 Mini            & $\mathbf{0.904} \pm 0.091$ & $0.882 \pm 0.095$ & $0.205$ \\
            \bottomrule
        \end{tabular}
    
\end{table}

\begin{table}[!b]
        \centering
        \renewcommand{\arraystretch}{1.2}
                \caption{Average number of iterations required per model (Mean $\pm$ Std. Dev.). Lower values indicate higher generation stability.}
        \label{tab:gen_iterations}
        \begin{tabular}{lcc}
            \toprule
            \textbf{Model} & \textbf{Baseline} & \textbf{Ours} \\
            \midrule
            Claude Haiku 4.5      & $1.60 \pm 1.29$ & $1.60 \pm 1.03$ \\
            Claude Sonnet 4.5     & $1.00 \pm 0.00$ & $1.00 \pm 0.00$ \\
            DeepSeek-v3.2         & $1.27 \pm 0.45$ & $1.60 \pm 0.92$ \\
            GPT 5.2               & $1.00 \pm 0.00$ & $1.00 \pm 0.00$ \\
            GPT-5 Mini            & $1.20 \pm 0.40$ & $1.00 \pm 0.00$ \\
            Gemini-3 Flash        & $1.20 \pm 0.40$ & $1.30 \pm 0.46$ \\
            Grok-4 Fast Reasoning & $1.00 \pm 0.00$ & $1.20 \pm 0.40$ \\
            Kimi K2               & $1.50 \pm 0.50$ & $\mathbf{2.86 \pm 1.89}$ \\
            Qwen3 Next 80B        & $2.75 \pm 1.54$ & $2.57 \pm 1.41$ \\
            \bottomrule
        \end{tabular}
    
\end{table}

\subsubsection{(ii) Qualitative Assessment for the Resource Perspective}
To validate the organizational perspective, we assess how accurately the LLMs identify and assign resource elements to process activities. Since exact string matching is insufficient given linguistic variability, we compare against the extended ground truth using two complementary scores:
\begin{description}
    \item[Semantic Similarity Score:] To account for valid paraphrases and abbreviations (e.g., ``HR'' vs.\ ``Human Resources''), we compute a semantic similarity score between generated and ground-truth labels using a pre-trained sentence embedding model~\cite{ReimersGurevych2021MiniLM} (\texttt{all-MiniLM-L6-v2}). We align predictions to ground truth \emph{by activity name}: for every activity $a$ appearing in both mappings, we embed $\textit{pool}_{gen}(a)$ and $\textit{pool}_{gt}(a)$ and compute their cosine similarity; the same procedure is applied to $\textit{lane}_{gen}(a)$ and $\textit{lane}_{gt}(a)$. Unspecified pools/lanes are represented as \texttt{None} and included in the comparison. Per-activity similarities are averaged into per-process \emph{Pool Similarity} and \emph{Lane Similarity} scores, then aggregated over all processes. Reporting pools and lanes separately disentangles coarse-grained actor identification from fine-grained role naming.
    \item[Judge-based Quality Score:] We implement an independent LLM-as-a-Judge protocol~\cite{DBLP:journals/corr/abs-2310-06271} with a fixed evaluator (\emph{Grok-4.1 Fast Reasoning}) separate from the generation models. For each instance, the judge receives the textual description and the activities with their predicted \texttt{pool}/\texttt{lane} assignments, but \emph{not} the ground-truth labels. It assigns a graded correctness score $s(a)\in[0,1]$ per activity, reflecting whether the assignment is supported by the textual evidence; minor naming variations are accepted, and \texttt{None}, i.e., no resource can be identified, is valid when the description provides no basis. The judge returns a strict JSON object $\{\textit{activity}\mapsto \textit{score}\}$; malformed or incomplete outputs trigger up to five retries before scoring proceeds. The per-process score is the mean of per-activity scores, aggregated across all processes for model-level results.
\end{description}

\subsection{Evaluation Results}

We now report the experimental results, examining first the impact on control-flow quality and efficiency, and then the accuracy of the discovered resource perspective.

\subsubsection{On the Task Complexity}
\paragraph{Impact on Control-Flow Quality:} Our experiments show that the control-flow quality scores for the resource-aware models closely mirror those of the baseline (cf. the aggregated results in \autoref{tab:f1_stats}). To rigorously validate this observation, we conduct significance testing on the control-flow quality scores using a paired t-test \cite{casella2024statistical}. The analysis reveals that for all cases the difference in performance between the control-flow-only and resource-aware models is not statistically significant ($p>0.05$). The obtained p-values range from 0.103 to 0.704, confirming that the additional complexity of generating pools and lanes does not degrade the underlying process logic.

\begin{figure*}[!t]
    \centering
    \includegraphics[width=0.86\textwidth]{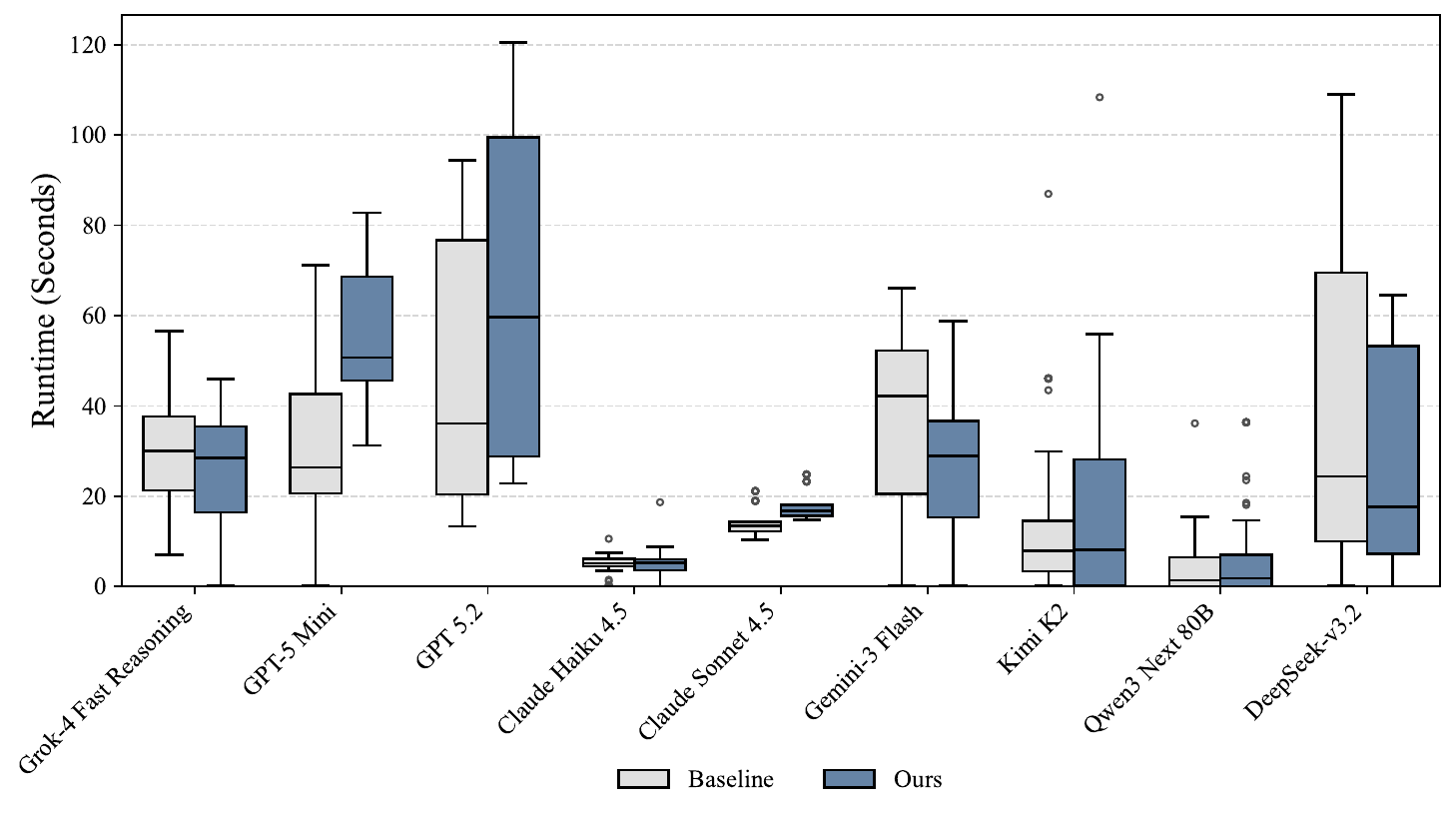}
    \caption{Boxplot analysis of execution time (seconds) per iteration. The consistent distributions between the baseline (grey) and our approach (blue) demonstrate the efficiency of the extended generation pipeline.}
    \label{fig:runtimes_iter}
\end{figure*}

\begin{figure*}[!t]
    \centering
    \includegraphics[width=0.86\textwidth]{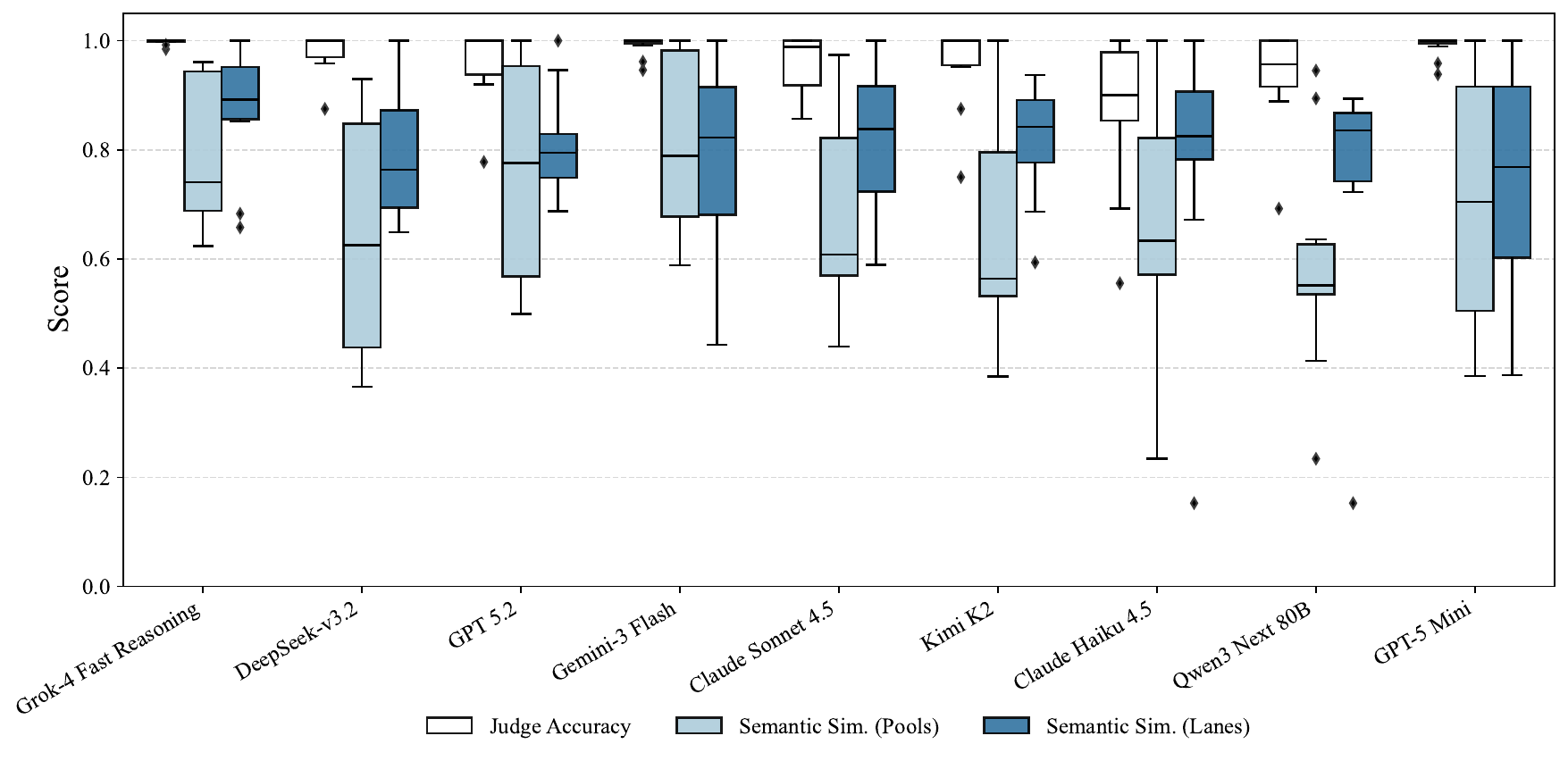}
    \caption{Accuracy of resource assignment aggregated by LLMs.}
    \label{fig:accuracy_metrics}
\end{figure*}

\paragraph{Efficiency Analysis}
\autoref{tab:gen_iterations} reports the number of generation iterations required until a valid model is produced. Most models remained virtually unchanged between both settings—Claude Sonnet 4.5 and GPT 5.2, for instance, achieved perfect scores ($1.00 \pm 0.00$) in both. The only notable exception is Kimi K2 (highlighted in bold), whose required iterations increased significantly from $1.50 \pm 0.5$ to $2.86 \pm 1.89$, suggesting that it struggles to simultaneously manage control-flow logic and resource assignment, leading to more syntactic or logical errors requiring correction.

\autoref{fig:runtimes_iter} compares per-iteration runtime distributions for the baseline and our resource-aware approach. Despite the longer generated code introduced by the pool and lane arguments, the runtime impact is negligible for most models: reasoning models like \emph{Grok-4 Fast} and \emph{Claude Sonnet 4.5} exhibit virtually identical profiles, while only larger models show a slight increase (e.g., GPT 5.2 from $\approx 40$ to $\approx 60$ seconds)—expected from the additional tokens and still well within acceptable limits for process modeling.

\subsubsection{Assessment of the Resource Perspective} 
\autoref{fig:accuracy_metrics} reports the semantic accuracy of the discovered pool and lane labels. The Judge-based accuracy (white bars) remains consistently high—top performers such as GPT-5 Mini achieve near-perfect scores across all processes—confirming that the discovered assignments are logically sound. Notable deviations appear, however, in the Pool Similarity (light blue bars), reflecting the lexical heterogeneity of process descriptions: a valid organizational entity approved by the Judge may still differ vectorially from the ground truth, as in ``Corporate Headquarters'' vs.\ ``Central Audit Team'' (p16) or ``Product Development Organization'' vs.\ ``Organization'' (p9). Some activities also inherently span multiple organizational boundaries (e.g., ``Sign Contract'' in p5, involving both Supplier and Procuring Entity), so either assignment is valid yet may be penalized by the similarity score. We therefore interpret Semantic Similarity as a measure of literal adherence to the ground-truth vocabulary, and Judge Accuracy as a more robust indicator of functional correctness.

\subsection{Findings}
Our results show a successful decoupling of modeling concerns: conceptualizing organizational boundaries does not impede the LLM's ability to maintain behavioral soundness in the control flow. Paired t-tests \autoref{tab:f1_stats} reveal no statistically significant difference in control-flow quality between the baseline and the resource-aware models ($p>0.05$) for any of the evaluated processes, indicating that identifying pools and lanes does not interfere with inferring the correct sequencing of activities. This semantic enrichment also incurs negligible computational overhead, with runtime distributions showing only a marginal increase for some models.

Regarding the resource perspective, the consistently high Judge-based accuracy scores confirm that LLMs possess robust semantic reasoning for identifying process participants. The lower semantic similarity scores in certain cases reflect lexical divergence from the ground truth, often representing valid alternative modeling decisions, though they may also stem from inherent model biases, similarly to what \cite{DBLP:journals/corr/abs-2509-15336} observe. Such ambiguity is intrinsic to textual descriptions, since a single text can admit multiple valid structures—motivating our use of a Judge-LLM metric, as deterministic measures cannot reliably credit semantically defensible alternatives. Finally, although generation is non-deterministic, prior benchmarking on this dataset~\cite{DBLP:conf/bpmds/KouraniB0A24} shows stable average performance across model families, supporting the reliability of the reported trends.

\section{Conclusion}
\label{chap:conclusion}
This paper moves generative process modeling beyond the control-flow perspective by integrating the resource dimension as a core component of the modeling process. By extending the POWL metamodel and leveraging a single-stage generation strategy, we demonstrate that LLMs can effectively capture the interplay between process logic and organizational boundaries. Our approach ensures the semantic coupling of these perspectives while offloading the complexities of formal serialization and layout to deterministic algorithms.

Evaluation across nine state-of-the-art LLMs confirms that adding organizational complexity does not degrade the behavioral soundness of the resulting models. The findings show that LLMs possess high semantic fidelity in participant discovery, maintaining practical efficiency with only marginal overhead compared to resource-agnostic baselines. By maintaining a deliberately simple architecture, we establish a robust foundation for future extensions-including data objects and temporal constraints—enabling the automated generation of high-fidelity, multi-collaborative process models.

\bibliographystyle{splncs04}
\bibliography{bibliography}

\end{document}